\documentclass[3p,times,procedia]{elsarticle}

\usepackage{ecrc}
\usepackage{graphicx}
\usepackage{amsmath,amssymb}
\usepackage{algorithm}
\usepackage{algpseudocode}
\usepackage[caption=false,font=footnotesize]{subfig}
\usepackage{graphicx}
\usepackage{textcomp}
\usepackage{url}
\usepackage{filecontents}
\usepackage{balance}

\volume{00}
\firstpage{1}
\journalname{Robotics and Autonomous Systems}
\runauth{K. Krinkin, A. Filatov}
\jid{procs}

\begin{document}

\begin{frontmatter}



\dochead{Robotics and Autonomous Systems}
%% e.g. \dochead{17th International Conference on Dynamical Processes in Excited States of Solids}

\title{Correlation Filter of 2D Laser Scans For Indoor Environment}


\author[a]{Kirill Krinkin}
\ead{kirill@krinkin.com}
\author[a]{Anton Filatov  \corref{cor1}}
\ead{ant.filatov@gmail.com}

\address[a]{Saint Petersburg Electrotechnical University "LETI", Saint Petersburg, Russia}

\begin{abstract}
Modern laser SLAM (simultaneous localization and mapping) and structure from motion algorithms face the problem of processing redundant data. Even if a sensor does not move, it still continues to capture scans that should be processed. This paper presents the novel filter that allows droping 2D scans that bring no new information to the system. Experiments on MIT and TUM datasets show that it is possible to drop more than half of the scans. Moreover thepaper describes the formulas that enable filter adaptation to a particular robot with known speed and characteristics of lidar. In addition, the indoor corridor detector is introduced that also can be applied to any specific shape of a corridor and sensor.
\end{abstract}

\cortext[cor1]{Corresponding author.}
\end{frontmatter}

%\correspondingauthor[*]{Corresponding author. Tel.: +0-000-000-0000 ; fax: +0-000-000-0000.}
%\ead{ant.filatov@gmail.com}

% \linenumbers
%\vspace*{-10pt}


%\enlargethispage{-7mm}

\section {Introduction}

Nowadays there are various application of lidars, laser scanners, rangefinders. Algorithms that use these sensors, such as SLAM, structure from motion and others, are highly demanded in modern robotics. Such sensors have a common disadvantage - they collect simultaneously too few and too much data. On the one hand there is too few data because it is impossible to smooth or approximate this data without significant loss of accuracy. At the same time there is too much data because it requires a lot of memory to store and process scans that appear every 30 milliseconds.

This paper presents an algorithm of filtering 2D laser scans in application to indoor SLAM algorithm. The development is relevant because modern laser scanners collect data more than 30 times per second. There is no need to capture laser scans so frequently unless such laser scanner is mounted on a car moving with 60 km/h. In this case the capturing environment might extremely change in 0.03 sec. On the contrary, if a robot moves in indoor environment and it has an average speed of about 0.5-1 m/s such amount of dense point clouds from laser rangefinder is excess.

The suggested algorithm is applicable to indoor environments. Moreover, it is useful in situation when a robot that processes lidar data has limited computational resources. In this case, even if the robot stays at one place, it should not process every successive scan and therefore it can save its resources. Also, if a robot moves slowly and smoothly, it is also possible to drop the scans that bring no new information.

The idea is based on storing several successive scans in a window and comparing each upcoming scan to the scans from this window. If that scan strongly correlates to every scan from the window - it should be dropped. Experimental testing on MIT Stata Dataset~\cite{fallon2013stata} shows that it is possible to drop more than half of the scans for needs of SLAM algorithm without loss of accuracy.

Another important feature of this algorithm is that the process of filtering takes less calculation time than scan matching in SLAM. Otherwise, it would be redundant to use this filtering. Every SLAM algorithm has its own unique scan matcher and it was impossible to compete with every one of them so the reference value of scan matching complexity was set to $O(100*points\_in\_scan)$ iterations. In average, a scan matcher that processes scans in real time should work with approximately this speed. The suggested filtering method works significantly faster than that and it is proved both mathematically and on experiments.

The limitations of the suggested algorithm are the following. The algorithm is assumed to be launched in indoor environment. The robot should be equipped with a lidar than captures laser scans more than 30 times per second. The scan capturing rate should be known. The robot should move no faster than on several centimeters per scan capturing. The average robot speed should be known. The robot can run any 2D SLAM algorithm, since the filtering process is inserted before SLAM activity.

The paper structure is as follows: the known approaches for scan filtration and calculation of scan correlation are described  in section 2. The description of the suggested method and it’s modification for featureless corridors is presented in section 3. Experimental testing on a real data can be found in section 4.

\section {Related work}
The idea of filtering scans presented in this paper comes from the problem of reducing computation time for scan matching. The common problem is to reduce the amount of data that should be processed instead of iterating through every point of a laser scan. There are several well-known techniques for reducing dimension of a laser scan to decrease the time of scan analysis. On the one hand, it is possible to drop some parts of scans or even the whole scans if they for some reason bring no new information. On the other hand, a raw scan can be transformed into another more lightweight form that whould be processed faster. 

Authors of~\cite{nieto2007recursive},~\cite{he2017iterative},~\cite{theiler2014keypoint} suggest extracting feature points from laser scans. This technique significantly decreases the dimension of input data. So further features might be considered, for instance, in Kalman filter. The disadvantage of such method is that the process of feature detection might take valuable time. Also, descriptors of such features should be robust to rotation and shift which is hard to achieve.

Another technique for reducing dimension is construction of histogram of a scan. For instance it was presented in~\cite{rofer2002using},~\cite{bosse2007histogram}. In~\cite{bosse2007histogram} this technique was used as a core feature of a scan matcher is SLAM algorithm. A histogram is more flexible than a set of features since it is easy to tune the size of a histogram, length of columns etc. However, it requires accuracy in these parameters or otherwise a histogram would not work.

After the dimension of input data is reduced it is required to detect whether a new piece of data is the same as the previous one or not. The easiest way appears to be a scan matching, as in~\cite{nieto2007recursive} or~\cite{diosi2005laser}. But the focus of this paper is on avoidong redundant matching if it is possible. Even for little dimensions of data scan matching process might take unreasonably much time.

Another known idea is to calculate correlation of histograms as it is described in~\cite{rofer2002using}. In this approach the histogram is considered as a random variable with statistical characteristics and therefore it is possible to apply classic approaches of Pearson, Kendall and others~\cite{bonett2000sample}.

Authors of~\cite{Lehtola2019preregistration} present the approach that is based on the similar mathematical apparatus. However their focus is to classify lidar data using spatial correlations. The approach that is described in that paper can be successfully applied to detect featureless areas of a single laser scan. However the focus of the suggested algorithm is to decide if the whole scan should be processed or not.

Also, there exists a completely different approach to estimate laser scan closeness to one another. The idea is based on loop closure from graph-based SLAM algorithm. There are various well-known works in this area, such as~\cite{hess2016real},~\cite{newman2006outdoor},~\cite{granstrom2009learning}. However, these approaches usually require more computation time than they can save. And their improvement in general requires reducing dimension of input data as well.

To sum up, the problem of making filtering faster than scan matching leads to the idea that it is necessary to reduce the amount of input data. After that, it is necessary to find out if the upcoming scan is valuable or not. Trying to extract more details from the scan would inevitably increase the calculation time. By details the specific parts of a scan or a shift between a current scan and a previous one are ment. If a scan is valuable, the filter should transfer an original scan to a core of a SLAM algorithm.

\section {Correlation filter for 2D laser scans}
The core idea of the suggested algorithm is to compare the current incoming laser scan to the previous one. If the latter is 'similar' than the current scan should not be processed. To avoid the noise in observations it is better to compare an incoming scan to several previous scans. Hence, a sliding window of scans appears that plays a role of reference for new incoming scans.

\subsection {Laser scan representation}
In general, a laser scan consists of several thousands points. The brute force calculation of scans correlation takes $O(n^2)$ operations, that is greater than a million. To decrease this dimension it is possible to extract feature points, as was described in section 2. However, this might still take enormous amount of time.

Hence, instead of using raw laser scan data for calculating correlation, it is suggested to create a histogram for each scan. There are several approaches for creating a histogram from a laser scan. One of them is based on a division by ranges, and another one - by angles.

For each scan, the highest and the lowest values of a range are known. Therefore, it is possible to divide this range dispersion in several intervals, and then calculate the number of points that fits each interval. 

Two successive scans should not differ significantly, so the range histograms should be close to each another. In practice, if a robot does not rotate, the difference between two scans is insignificant and the values of each column of histogram varies in few units. If a robot rotates the difference is more considerable. However, it is possible to update the approach of range histogram. Instead of calculating the quantity of points in each column, it is possible to calculate the average range of column. Hence, two successive histograms become more diverse, allowing to avoid excess dropping of scans.

There is another approach to create a histogram that is opposite to a division by range. Every laser scan is captured in polar coordinates where an angle is the second degree of freedom in addition to a range. Thus, it is possible to divide every scan into several intervals of angles and calculate the average ranges in these intervals. It is impossible to calculate the number of ranges because it is the same for each interval. It is also possible to calculate dispersion instead of average value.

Both approaches for creating a histogram decrease the amount of calculations. Instead of processing thousand points of laser scan for further calculation of correlation, it is possible to process several dozens of columns in histograms. It is important to mention that a laser scan is not replaced with a histogram in SLAM algorithm. The histogram is created only to calculate the correlation of scans. 

In addition, experiments show that histograms correlation in every window are calculated faster than a scan matcher finds the best position for a scan. Therefore it is possible to filter excess scans and gain extra time resources for other routines. The specific numbers are presented in section 4.

\subsection{Criteria of correlation}
The next step after the histograms of each scan have been created, is to calculate their correlation. Methods of mathematical statistics can be applied here, considering the histogram of a scan as a random variable with unknown distribution. Since every histogram from a window in general is similar to each other, it is possible to assume that the distribution is the same.

There are several known approaches to calculate correlation of random variables: Pearson correlation coefficient~\cite{benesty2009pearson}, Kendall coefficient~\cite{abdi2007kendall} and Spearman coefficient~\cite{myers2004s}. Pearson coefficient is calculated to a pair of random variables $X$, $Y$ using the formula

$$P_{X,Y}=\frac{cov(X,Y)}{\sigma_X\sigma_Y}$$

where $cov$ - is the covariance, \\
$\sigma_X$ - is the standard deviation of $X$, \\
$\sigma_Y$ - is the standard deviation of $Y$.

If random variables are observed $n$ times and $x_i$ is the observation of variable $X$ than the Pearson correlation coefficient is calculated with the following formula

\begin{equation}
P_{X,Y}=\frac{n(\sum_{i=1}^{n}x_i y_i) - (\sum_{i=1}^{n}x_i)(\sum_{i=1}^{n}y_i)}{\sqrt{\begin{bmatrix}n\sum_{i=1}^{n}x_i^2-(\sum_{i=1}^{n}x_i)^2\end{bmatrix}\begin{bmatrix}n\sum_{i=1}^{n}y_i^2-(\sum_{i=1}^{n}y_i)^2\end{bmatrix}}}
\end{equation}

The value of this coefficient is between -1 and 1. A value of +1 means total positive linear correlation, 0 means no linear correlation, and -1 means total negative linear correlation. This correlation is called linear because of the following interpretation. The value of the first variable is put in abscissa of a plot, the value of the second variable - in ordinate. If the points with resulting coordinates belong to one line with positive derivative, then their correlation is equal to 1. If the derivative is negative, then the Pearson coefficient is equal to -1. If there is no way to draw any line - then the coefficient is 0.

Kendall and Spearman coefficients are used to measure the ordinal association between two measured quantities. It is a measure of rank correlation: the similarity of the data orderings when ranked by each of the quantities. The Spearman coefficient is defined as the Pearson coefficient between the rank variables. In other words, if $X$ and $Y$ are random variables with ranks $rg_X$, $rg_Y$ the correlation coefficient is calculated as

$$P_{rg_X, rg_Y} = \frac{cov(rg_X,rg_Y)}{\sigma_{rg_X}\sigma_{rg_Y}}$$

The calculation of Kendall coefficient is associated with the number of concordant pairs of random values. Let ($x_i$, $y_i$) be a set of observations of random variables X and Y. Pairs ($x_i$, $y_i$) and ($x_j$, $y_j$) where i < j are said to be concordant if either both $x_i\ >\ x_j$ and $y_i\ >\ y_j$ holds or both $x_i\ <\ x_j$ and $y_i\ <\ y_j$. Otherwise they are called discordant.
The Kendall tau coefficient is defined as

$$P_{X,Y} = \frac{number\ of\ concordant\ pairs - number\ of\ discordant\ pairs}{\binom{n}{2}}$$

To sum up, there are three well known approaches to calculate correlation. The main disadvantage of Kendall coefficient is the algorithmic complexity. It requires calculating the rank of random variable and then calculating the number of concordant pairs. In the worst case this might take $Nlog(N)$ operations. Spearman coefficient is more lightweight, but it also requires ranking. Since the correlation is calculated for histograms of successive scans, it is a challenging problem to rank values in histograms in the correct way. This means that the histograms are, in general, similar, and it is necessary to capture every little fluctuation of values. Therefore, the rank function should be sensitive for these fluctuations and simultaneously be equitable.

Therefore, the Pearson correlation coefficient is the most suitable for the considered algorithm. Its complexity is $O(n)$, it does not require ranking function, and it is sensitive enough to fluctuations of values in histograms.

\subsection{Parameters and constants}
At the moment, there are four places, where parameters adjustment are to be considered:
\begin{itemize}
    \item amount of columns in histogram - and also amount of points in each column;
    \item size of a window;
    \item threshold of intra window correlation - how much a new upcoming scan should be correlated to each scan in window. It is indicated below by $P_{pair}$;
    \item dropping correlation threshold - combined value of scan similarity to scans in a window ($P_{common}$);
\end{itemize}

These four parameters influence the amount and nature of dropped scans. First of them is the amount of columns in histogram, that was described in 3.1. All considered histograms have a common feature: the more columns are contained in a histogram, the more details of each scan are processed. Consider lidar used in MIT dataset, that captures scans consisting of approximately 1000 points. Then the division of these points in 50 columns means that each group of points contains in average 20 points. The division in 10 columns brings the groups with 100 points in each. 

To determine the influence of columns number on the quantity of information, that is possible to be received from histogram two border situations can be considered. First situation implies, that each point of scan is contained in a separate column. Therefore, the histogram consists of 1000 columns. Correlation between two such histograms is very sensitive to every point and even noise, but it can show the true similarity of laser scans. On the other hand all can be gpoints in one column. In this case, all scans become similar, and there appears no way to distinguish one from another. 

This example shows, that the larger amount of columns leads to the larger sensitivity of scan correlation to a minor difference of scans. Despite the intuitive opinion, that the higher sensitivity leads to the higher accuracy, in real data the high sensitivity might be, on the contrary, unpleasant. For instance, if a moving object appears in the area of laser scan view, it inevitably brings difference in two successive scans. Moreover every sensor brings noise in observations and sometimes this noise can be falsely interpreted for little distances as a difference between scans. Experiments results presented in section 4, show that 1000 points in laser scan, that are spread almost in the field of view of $3\pi/4$, should be grouped in 15 or 30 columns.

Another important constant is the size of the window, that contains previous laser scans. The score of a current scan is equal to the correlation value  multiplication of each scan from this window to a current scan. By correlation value the Pearson correlation coefficient is assumed. It is obvious, that if a robot with a laser scanner moves quickly, then large amount of scans in the window brings little final correlation coefficient. That means, that the higher robot’s speed the lower number of scans in the window is appropriate.

Experimentally the formula appeared that links a windows size to an average speed of robot. Under the average speed here, the average distance in centimeters, that a robot passes between two scan captures, is assumed. This formula is heuristic and allows to link the property of scan capturing - speed - and the property of filter - amount of information, that should be in cache.

\begin{equation}
window\_size = \frac{27}{avg\_speed^2}
\end{equation}

In order to clarify the influence of a window size it is necessary to determine two parameters, that are strongly related to each other and to a window size. The first parameter is the threshold for Pearson correlation coefficient of each pair of scans, and the second parameter is the common correlation coefficient, that is equal to the multiplication of coefficients. Since the Pearson correlation coefficient is calculated for two successive scans, that are captured with a little time difference, it is obvious, that they are, in average, very strongly correlated. That is why a threshold for a pair of scans should be no lower than 0.95, or even better - 0.98. After calculating correlation coefficient of new scan to each scan in a window, it is necessary to combine all coefficients. The obvious and the well-known way to do that is to multiply them:
\begin{equation}
P_{common} = \prod_{i = 1}^{window\_size} P_{pair} = (P_{pair})^{window\_size}
\end{equation}

This formula allows estimating the common threshold according to a chosen window size and a pair correlation coefficient. For instance, a threshold for a window containing 5 scans and pair correlation 0.98 is equal to $0.98^5 = 0.904$.

\subsection{Corridor detector}
The general concept of the filtering method is based on the idea, that similar scans might be dropped. To achieve that, a histogram is built for each scan, and the correspondence of these histograms is calculated. If a robot stays at one place, it is obvious that it captures the same scans, and all of them except one might be dropped. But, if a robot moves, it is unlikely, that it receives the same scans. However, it is possible in featureless environments.

\begin{figure}[h]
	\centering
\includegraphics[width=0.8\textwidth]{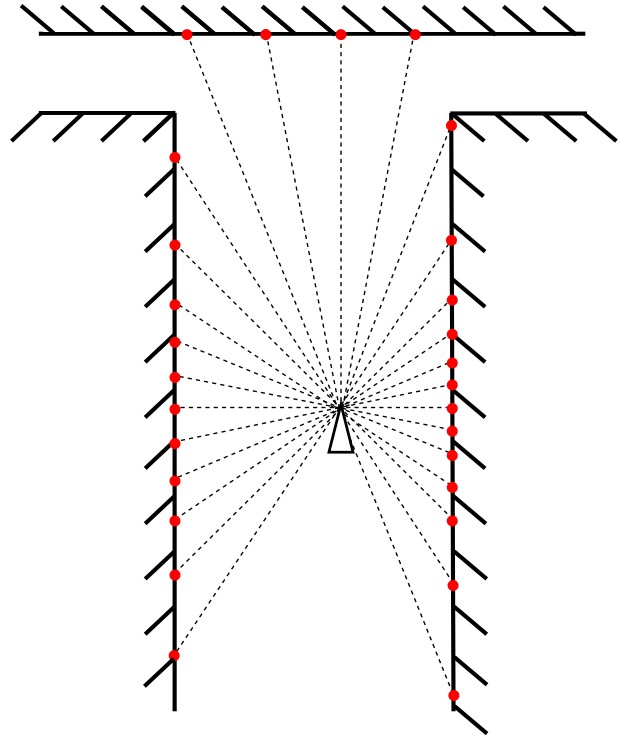}
	\caption{Corridor scan. The robot position is indicated by a triangle. The dotted lines mean the lasers of lidar}
	\label{figCorridor}
\end{figure}

In case of indoor environment, the outstanding featureless environment is a corridor. When a robot moves through a corridor along the walls, there are only a few points that differ in successive scans. The example of such corridor scan can be found in fig~\ref{figCorridor}. Obviously, corridors are the most challenging pieces of environment for scan matcher, and, therefore, scans of corridors should not be dropped at all. Otherwise, there is a probability to loose useful information.

\begin{figure}[h!]
	\centering
\includegraphics[width=0.8\textwidth]{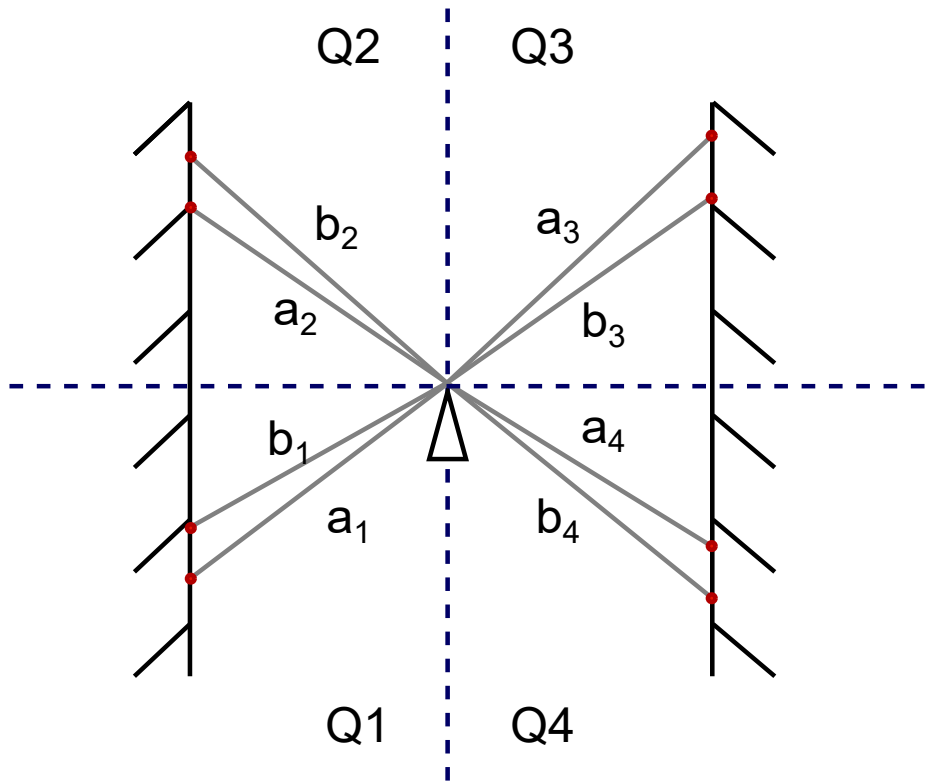}
	\caption{Corridor scan divided in quarters to demonstrate the relation of successive ranges}
	\label{figQuarters}
\end{figure}

For such scans, it is easy to expect, that their correlation is very close to 1 and they will inevitably be dropped. Therefore, it is required to detect corridor scans before filtering and not drop them. 
The suggested method is based on an assumption, that the corridor walls are straight and featureless. Hence, the ranges in a laser scan change monotonically. The easiest way to check this is to look through every point of a laser scan and to calculate the sign of range difference of the current point and the next one. It is clear, that it is necessary to consider such differences in every quarter of the field of view independently. The explanation of this fact is presented in fig~\ref{figQuarters}.

If a robot is placed along the corridor, the difference between $a_1$ and $b_1$ should be positive. At the same time, the difference between $a_2$ and $b_2$ should be negative, despite all these four points are located in the same wall. Also, the difference between $a_3$ and $b_3$ should be positive, and the difference between $a_4$ and $b_4$ - negative. If a robot is placed perpendicularly to the corridor, these signs are inverted, but their relation stays the same.

The general algorithm of corridor detection is presented below
\begin{algorithm}
\begin{algorithmic}

\State $score \gets 0$
\ForAll{$point \in laser\_scan$}
   \If{$point \in quarter_1\ \text{or}\ quarter_3$}
        \State $score \gets score + \text{sign}(point - point\_next)$
    \Else
        \State $score \gets score - \text{sign}(point - point\_next)$
    \EndIf
\EndFor
\State $score \gets score / \text{sizeof}(laser\_scan)$
     
\end{algorithmic}
\end{algorithm}

If the resulting score is greater than a threshold, then the corridor is found. The threshold depends on a restrictions of a concept of 'corridor'. If a corridor is a completely featureless pair of infinite plain walls, then this threshold might be close to 1. If a wall of a corridor might contain roughness, or something might appear in the middle of a corridor, then this threshold should be decreased. Experiments on a real data show, that the threshold equal to 0.5 makes it posible to detect corridors without false negative errors. There might appear false positive detections of a corridor, but in MIT dataset they were lower than 0.1\% of such detections in each sequence.

\begin{figure*}[h!]
	\centering
	\subfloat[][the angle between beams is big enough to keep a positive angle between $x+\Delta x$ and $y-\Delta y$ \\ ]{\includegraphics[height=0.31\textheight]{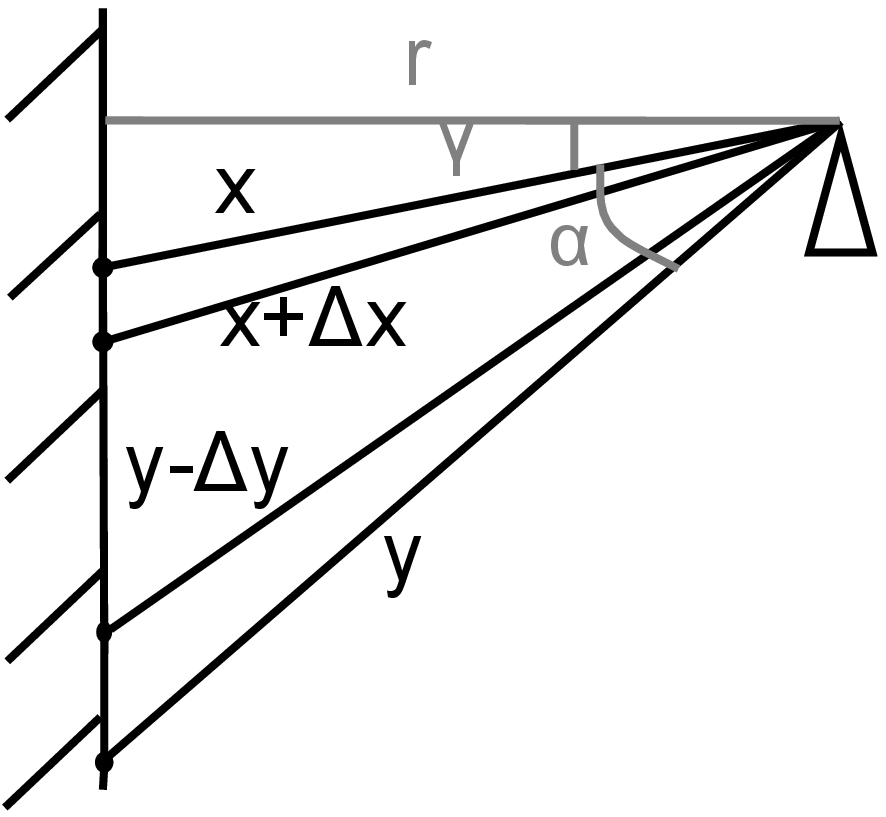}}
	\hfill
    \subfloat[][border case \\ ]{\includegraphics[height=0.31\textheight]{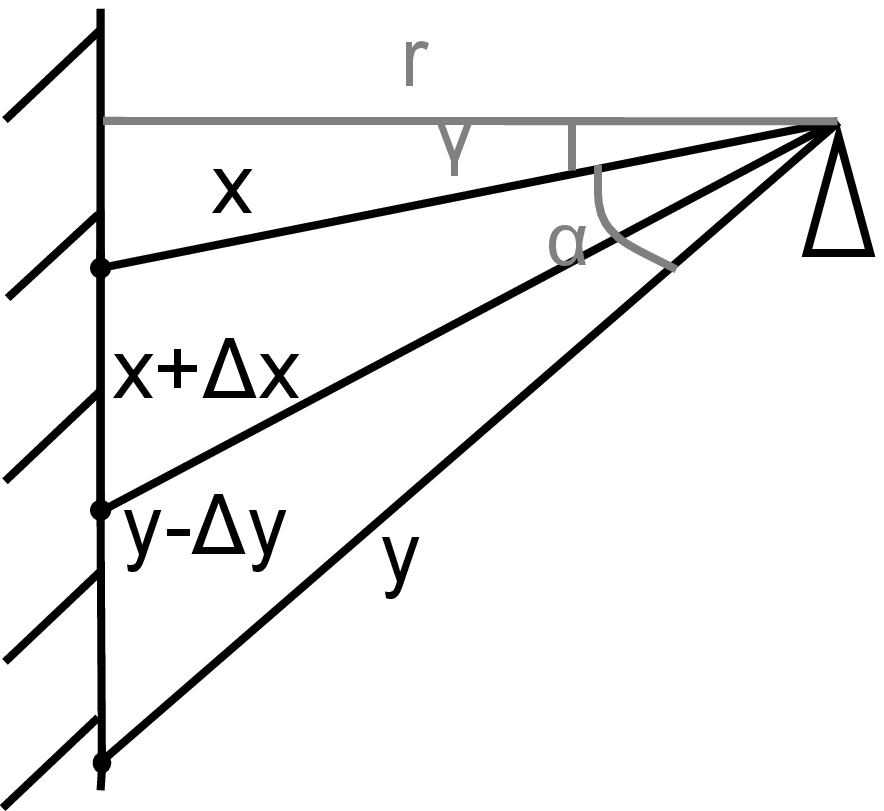}}
    \hfil
    \caption{Illustration of successive ranges x and y and their errors}
	\label{figGeometry}
\end{figure*}

The last but not least thing to discuss is the point\_next in the algorithm above. The obvious way is to take the truly next point to the current one. Nevertheless, this choice fails if the ranges of successive points are close to each other and differ no greater than the error of a laser scanner. A problem appears of detecting the minimal angle between laser beams hitting one wall. The laser beams contracting this angle should just have the ranges, that are not sensitive to a laser scan error. Fig~\ref{figGeometry} presents the geometrical illustration of this task.

Fig~\ref{figGeometry} presents two laser beams with ranges $x$ and $y$. The angle $\alpha$ is unknown, but it should be enough to distinguish x and y including their error. The errors are named $\Delta x$ and $\Delta y$, but in general they should be equal. If $\alpha$ is high enough, then the angle between $x+\Delta x$ and $y-\Delta y$ is positive as it is presented in fig~\ref{figGeometry}a. Decreasing $\alpha$ leads to decreasing the angle between beams with errors up to the border case presented in fig~\ref{figGeometry}b. In these figures $r$ - is a distance from scanner to the corridor; $\gamma$ - an angle between current beam and the edge of a quarter.
As fig~\ref{figGeometry}b shows, the border case means that
$$x + \Delta x = y - \Delta y$$
or
$$y - x = 2\Delta$$
Expressing $x$ and $y$ with $k$, alpha and gamma, the expression can be drawn:
$$\frac{r}{\cos(\gamma + \alpha)} - \frac{r}{\cos(\gamma)} = 2\Delta$$
Hence,
$$\cos(\gamma + \alpha) = \frac{r\cos(\gamma)}{r + 2\Delta \cos(\gamma)}$$

The same expression might be considered for each quarter of a laser scan, and they will be the same except of the sign of fraction. Therefore, it can be assumed, that gamma varies from 0 to $\pi/2$, where the cosine function is monotonous. So, it is enough to consider gamma, that is equal to 0 and $\pi/2$, to find out the border value for alpha. Hence, the expression is obtained, showind that the closer scan point is to the perpendicular to the corridor, the stronger is the range sensitivity to scanner error.

$$\cos(\alpha)=\left\{\begin{matrix}
r(r+2\Delta)^{-1}, &\text{if }\quad \gamma = 0
\\ 
0, & \quad \text{if} \quad \gamma = \pi/2
\end{matrix}\right.
$$

To sum up, the point\_next should be as far from current point as the following requirement is satisfied: $\cos(\alpha) = r(r+2\Delta)^{-1}$, where $r$ - is the range to a wall of corridor, and $\Delta$ - is a possible noise of such observation. These parameters can be obtained from the characteristics of a particular lidar.

\section{Evaluation}
In order to test the scan filter, it was inserted in three SLAM algorithms: vinySLAM~\cite{huletski2017vinyslam}, Gmapping~\cite{grisettiyz2005improving} and Google~Cartographer~\cite{hess2016real}. The filter decides if the scan should be processed or dropped before it is passed to a scan matcher. Hence, if a scan should be processed, then the general processing time consists of both filtering time and scan matching time. That is why it is necessary to estimate the complexity of filtering process. The first part of this paragraph is focused on this problem. The second part presents the results of such filtering application on real datasets, such as MIT stata dataset and TUM dataset. It is necessary to estimate the percentage of scans, that might be dropped without loss of accuracy. These datasets were chosen, because they contain laser scan data and are provided with arbitrary ground truth.

It is important to mention that the effect of corridor detector cannot be measured quantitatively. It is only possible to say that without this algorithm almost every laser scan of a corridor is dropped, and this dramatically affects on the trajectory estimation. In other words, the robot gets lost in the corridors. When the filter is included the corridor scans are not dropped and the robot performs ordinary SLAM.

\subsection{Filter algorithm complexity}
\begin{figure}[h]
	\centering
\includegraphics[width=1.0\textwidth]{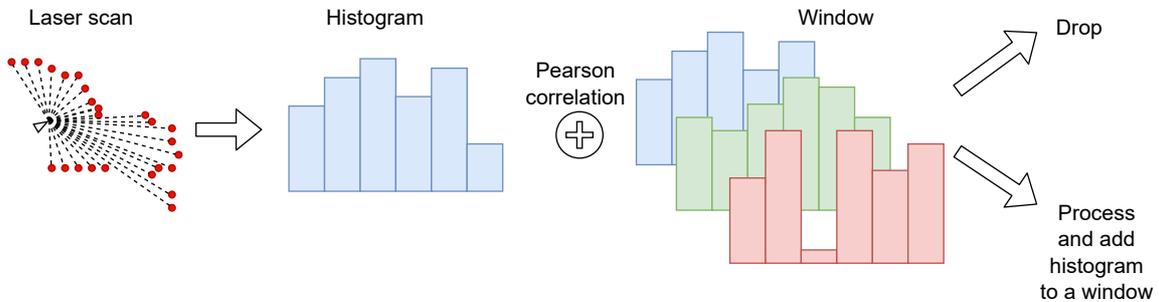}
	\caption{The scheme of communication in suggested algorithm}
	\label{figAlgorithm}
\end{figure}

The full process of filtering is presented in fig~\ref{figAlgorithm}.

First, it is necessary to create a histogram by looking through every point of laser scan. This takes $O(n)$ operations, where $n$ - is the number of points. To be more specific, a histogram, that shows the average range for each angle interval, takes $n$ multiplications, $n$ additions and $1$ division. Therefore, the process of histogram creation takes roughly $2n$ operations, without considering memory costs. The second step - the calculation of Pearson correlation according to the formula (1) - takes $5m$ additions and $3m$ multiplications, where $m$ is the number of columns in histogram, which is usually more than ten times lower, than $n$. Since Pearson correlation should be calculated for each scan in a window, it takes $8mk$ operations where $k$ is the size of a window.

To sum up, a very rough estimation, that considers only the most important parts of the algorithm, says that there are about $2n+8mk$ operations. In the real experiment $n$ is about 1000, $m$ is about 30-50 and $k$ is 5-10. Using this estimation, it is possible to say, that filtering process takes the same time as several iterations through all points of laser scan. At the same time, the scan matching process in vinySLAM represents almost random walk, that takes about 100 iterations through all points of laser scan. So, a rough estimation of scan matching is $100n$ operations. These estimations are not accurate, but they demonstrate, that the time for filtering should be significantly lower, than the time for scan matching.

\begin{figure}[h]
	\centering
\includegraphics[width=1.0\textwidth]{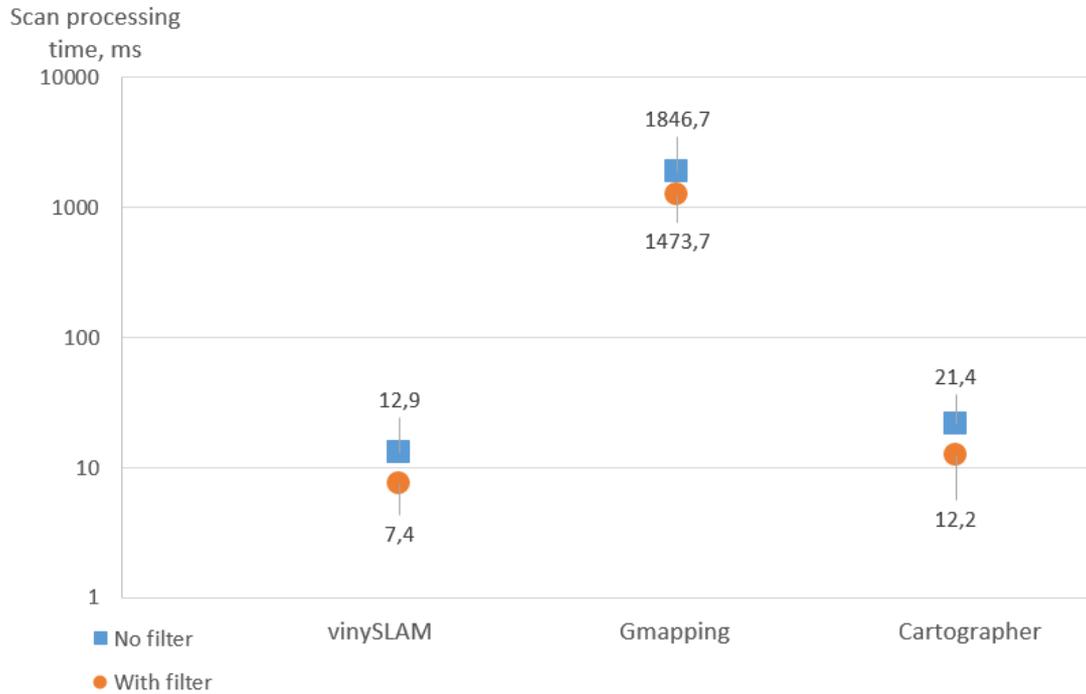}
	\caption{Scan processing time with and without filter for each considered SLAM algorithm (Logarithmic scale)}
	\label{figPerformance}
\end{figure}

It is also necessary to measure the real difference between scan matching time and filtering time. The experiment was made using the sequence mit-2011-01-25-06-29-26 on Ubuntu 18.04, ROS melodic, Intel Core I5-8500@ 3GHz, 16 Gb RAM. The average scan processing time for this sequence is measured for considered algorithms. Firstly, the time was measured without filtering. The result can be found in figure~\ref{figPerformance}. In this figure the average processing time of scan with filtering is also presented. The result shows that the mean scan processing time is decreased by more than 40\%. The average scan processing time in vinySLAM algorithm is $12.9\cdot10^{-3}$ seconds, and the average filtering time for 30 columns in histogram and 5 elements in window is $5.9\cdot10^{-5}$ seconds.

\subsection{Quantitative estimation of the filter}

The same hardware environment was used to calculate the accuracy of SLAM algorithms with filtering on MIT and TUM datasets. These datasets were chosen, because they are provided with ground truth, which allows estimation of the accuracy quantitatively. The test case is as follows:

\begin{enumerate}
\item Run pure SLAM algorithm on the sequences of these datasets, and measure its RMSE from the ground truth.
\item Run SLAM algorithm with filtering and estimate the accuracy with some dropped scans from the data sequences.
\item Estimate the percentage of dropped scans.
\end{enumerate}

The results on MIT dataset are presented in tables~\ref{tabMIT1} and~\ref{tabMIT2}. It is clear, that the accuracy of every SLAM with filter is the same as the accuracy of this algorithm without filter. At the same time, more than half of the scans is dropped. It is important to mention, that parameters for the first 7 sequences differ from parameters for the last 3. They are grouped by the average speed of robot.

The average speed of robot in first group is 0.022 m/quantum. Quantum is a period of time between capturing scans. For this particular laser scanner, it is equal to 0.025 sec.
According to a formula (2), the window size should be equal to 5. At the same time, the correlation coefficient for the pair of scans is 0.96, since the robot moves fast. Therefore, according to formula (3), $P_{common}$  is equal to 0.8. The value of $P_pair$ is heuristic and makes $P_common$ high enough. The last parameter to settle is the nu,ber of columns, which is also heuristic. It is strongly related to $P_pair$. It should be 30 to make the histogram sensitive to possible changes of environment.
For the last sequences, the average speed of robot is close to 0.017 m/quantum. This speed is lower, and therefore, window size is equal to 9, $P_{pair}$ = 0.99, $P_{common}$=0.9, the number of columns is equal to 15.

\begin{table}[h!]
\centering
\caption{RMSE values and percentage of dropped scans for vinySLAM and Gmapping on MIT dataset}
\label{tabMIT1}
\begin{tabular}{|c|c|c|c|c|c|}
\hline
The sequence                &      vinySLAM            & vinySLAM filter    & Gmapping & Gmapping filter & \% dropped \\
\hline
2011-01-20-07-18-45     &      0.062 $\pm$ 0.004          &       0.078 $\pm$ 0.004     &      0.216 $\pm$ 0.019          &       0.222 $\pm$ 0.021 & 59 \\
\hline      
2011-01-21-09-01-36     &      0.080 $\pm$ 0.018          &       0.089 $\pm$ 0.013     &      0.228 $\pm$ 0.064          &       0.221 $\pm$ 0.069 & 56\\
\hline
2011-01-24-06-18-27     &      0.096 $\pm$ 0.007          &       0.111 $\pm$ 0.013     &      0.232 $\pm$ 0.014          &       0.247 $\pm$ 0.041 & 59\\
\hline
2011-01-25-06-29-26     &      0.094 $\pm$ 0.006          &       0.100 $\pm$ 0.002     &      0.204 $\pm$ 0.007          &       0.191 $\pm$ 0.013 & 63\\
\hline
2011-01-27-07-49-54     &      0.170 $\pm$ 0.019          &       0.121 $\pm$ 0.006     &      0.299 $\pm$ 0.032          &       0.309 $\pm$ 0.022 & 52\\
\hline
2011-03-11-06-48-23     &      0.534 $\pm$ 0.085          &       0.543 $\pm$ 0.034     &      0.574 $\pm$ 0.059          &       0.633 $\pm$ 0.046 & 58\\
\hline
2011-03-18-06-22-35     &      0.090 $\pm$ 0.020          &       0.090 $\pm$ 0.003     &      0.167 $\pm$ 0.013          &       0.149 $\pm$ 0.028 & 52\\
\hline
\hline
2011-04-06-07-04-17     &      0.183 $\pm$ 0.014          &       0.213 $\pm$ 0.027     &      0.328 $\pm$ 0.037          &       0.389 $\pm$ 0.054 & 51\\
\hline
2011-01-19-07-49-38     &      0.305 $\pm$ 0.174          &       0.289 $\pm$ 0.181     &      0.242 $\pm$ 0.010          &       0.212 $\pm$ 0.032 & 50\\
\hline
2011-01-28-06-37-23     &      0.361 $\pm$ 0.175          &       0.348 $\pm$ 0.152     &      0.371 $\pm$ 0.032          &       0.339 $\pm$ 0.039 & 47\\
\hline
\end{tabular}
\end{table}

\begin{table}[h!]
\centering
\caption{RMSE values and percentage of dropped scans for Google Cartographer on MIT dataset}
\label{tabMIT2}
\begin{tabular}{|c|c|c|c|}
\hline
The sequence                &      Google Cartographer            & Google Cartographer filter      & \% dropped\\
\hline
2011-01-20-07-18-45     &      0.131 $\pm$ 0.058          &       0.139 $\pm$ 0.041     &      59  \\
\hline      
2011-01-21-09-01-36     &      0.153 $\pm$ 0.072          &       0.163 $\pm$ 0.094     &      56 \\
\hline
2011-01-24-06-18-27     &      0.183 $\pm$ 0.015          &       0.181 $\pm$ 0.014     &      59 \\
\hline
2011-01-25-06-29-26     &      0.176 $\pm$ 0.010          &       0.179 $\pm$ 0.012     &      63 \\
\hline
2011-01-27-07-49-54     &      0.248 $\pm$ 0.014          &       0.251 $\pm$ 0.007     &      52 \\
\hline
2011-03-11-06-48-23     &      0.586 $\pm$ 0.174          &       0.642 $\pm$ 0.191     &      58 \\
\hline
2011-03-18-06-22-35     &      0.130 $\pm$ 0.025          &       0.119 $\pm$ 0.017     &      52 \\
\hline
\hline
2011-04-06-07-04-17     &      0.188 $\pm$ 0.011          &       0.185 $\pm$ 0.011     &      51 \\
\hline
2011-01-19-07-49-38     &      0.188 $\pm$ 0.004          &       0.189 $\pm$ 0.005     &      50 \\
\hline
2011-01-28-06-37-23     &      0.378 $\pm$ 0.025          &       0.399 $\pm$ 0.030     &      47 \\
\hline
\end{tabular}
\end{table}

The slight increase of ground truth quality on several sequences can be explained with statistical error, since all considered algorithms implement different variations of random walk in their scan matchers.

\begin{figure*}[h!]
	\centering
	\subfloat[][vinySLAM \\ ]{\includegraphics[height=0.34\textheight]{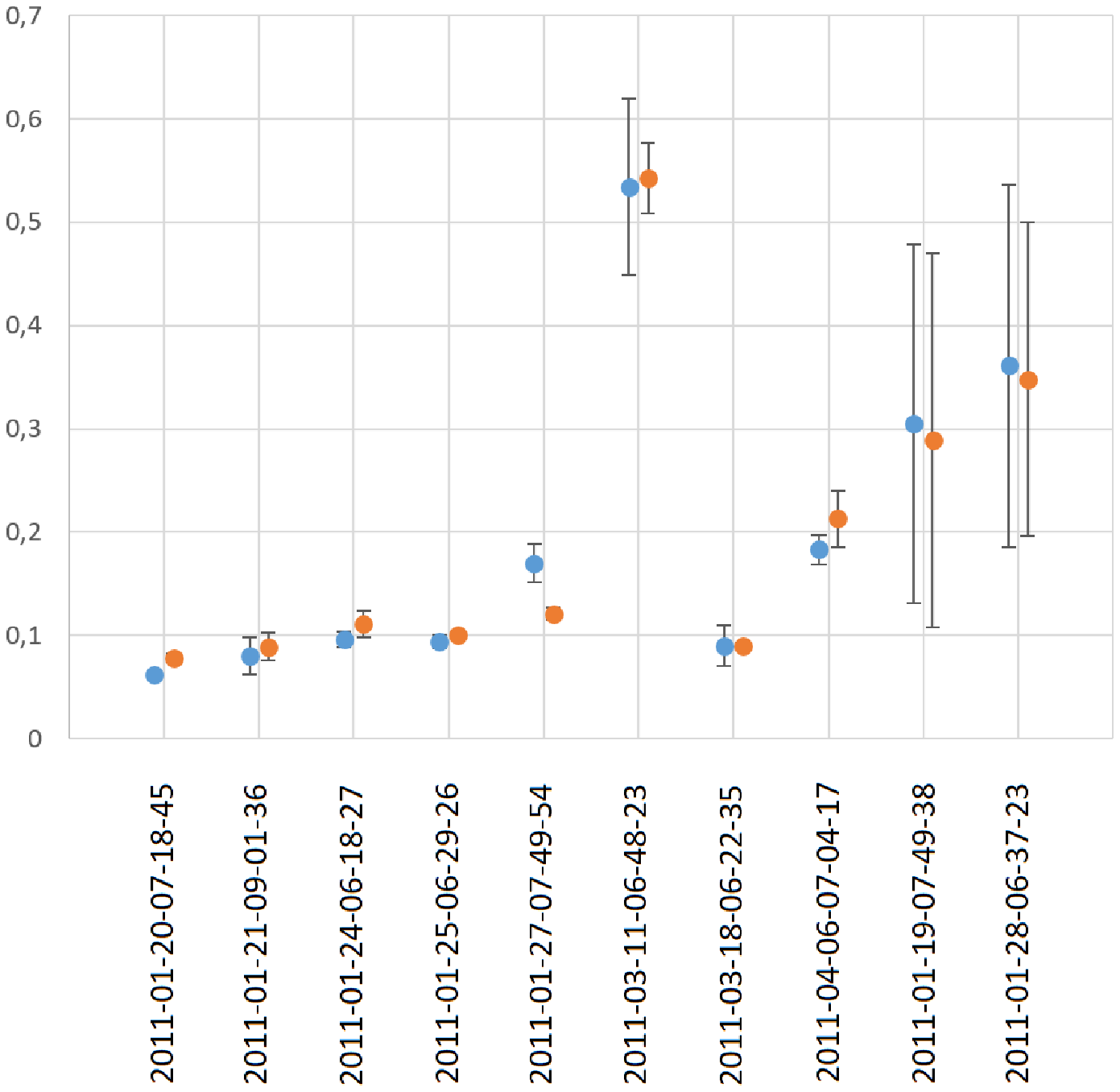}}
	\hfill
    \subfloat[][Gmapping \\ ]{\includegraphics[height=0.34\textheight]{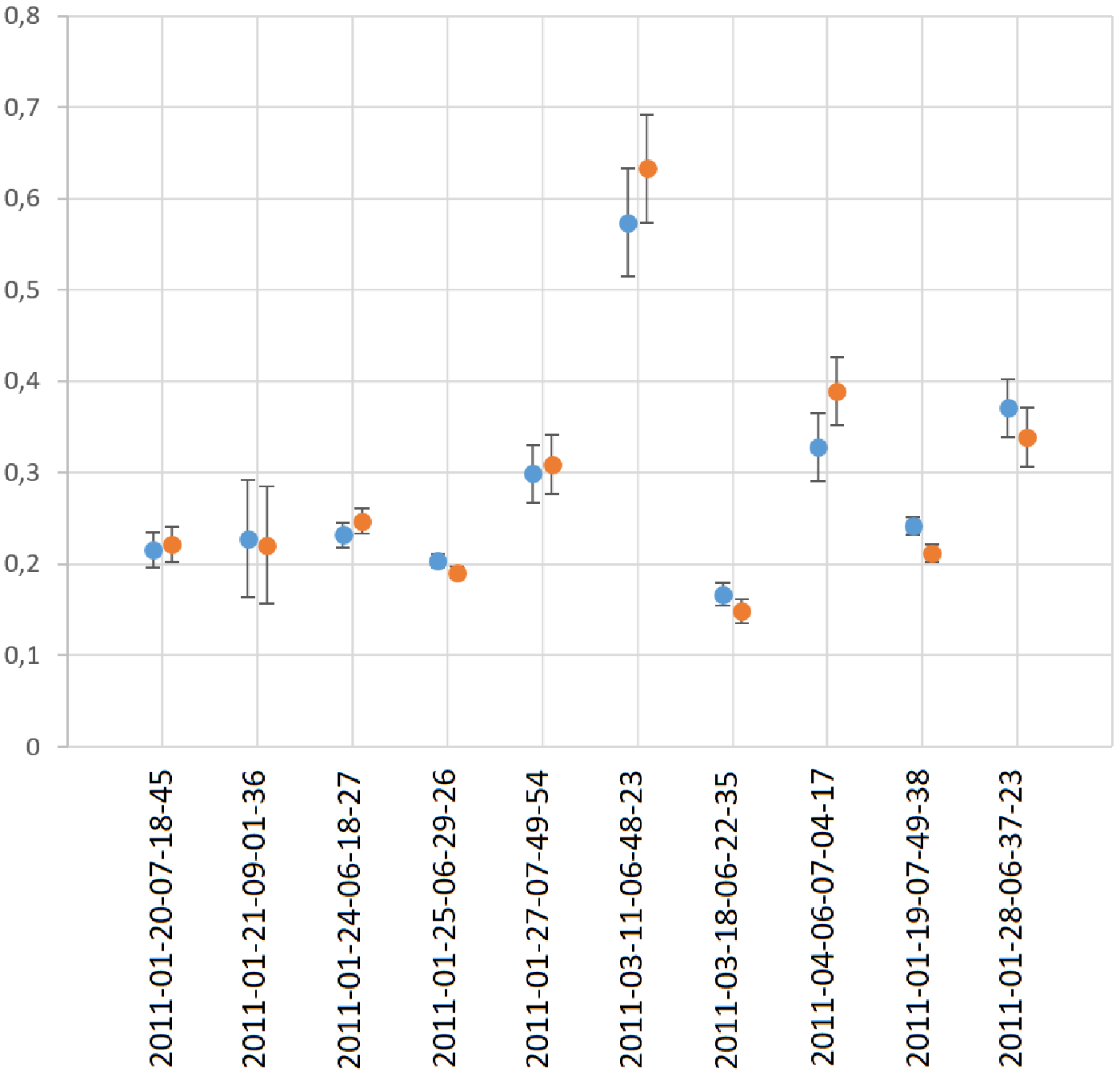}}
    \hfil
    \subfloat[][Cartographer \\ ]{\includegraphics[height=0.4\textheight]{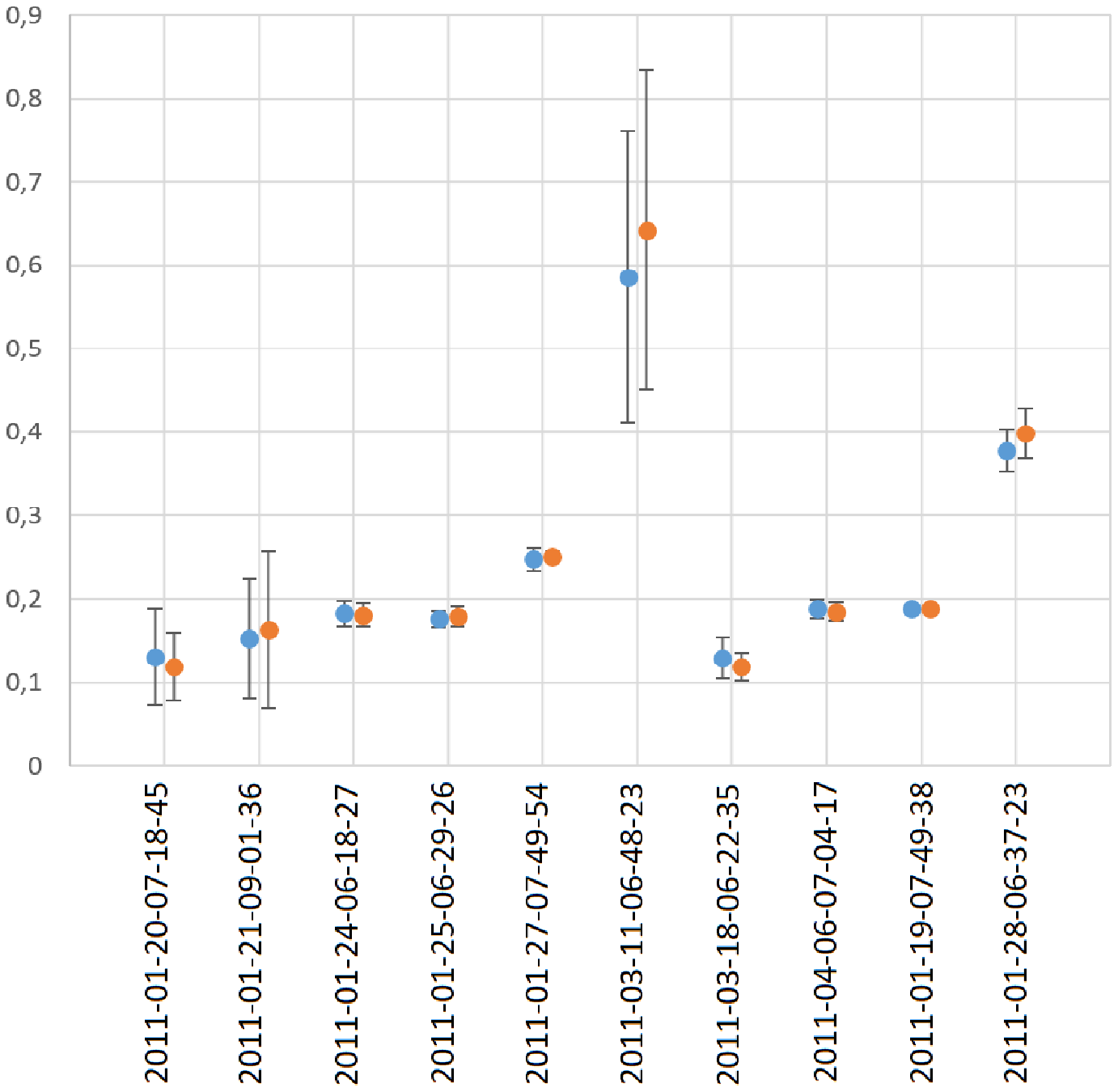}}
    \hfil
    \caption{Visualization of SLAM results with filter turned on and off. For each sequence the result without filter is on left and with filter is on right}
	\label{figSLAM}
\end{figure*}

The figure~\ref{figSLAM}a presents the results visualization of vinySLAM with filter and without filter. It is clear, that they are the same, taking the error into account. The figure~\ref{figSLAM}b shows these results for Gmapping and fig~\ref{figSLAM}c provides the visualization of tab~\ref{tabMIT2} (Google Cartographer)

The TUM dataset contains few sequences with laser scan, and therefore, the results are not as indicative as on MIT. The results of vinySLAM are presented in table~\ref{tabTUM}. The average speed is 0.024 m/quantum  for the first sequence and 0.021m/quantum for the second one.
They are launched on the following parameters:
$window\_size$ = 5, $P_{pair}$ = 0.98, $P_{common}=0.9$, $number\ of\ columns$ = 30.

The source code can be found at \url{https://github.com/OSLL/slam-constructor} (branch filter).

\begin{table}[h!]
\centering
\caption{RMSE values and percentage of dropped scans for vinySLAM on TUM dataset}
\label{tabTUM}
\begin{tabular}{|c|c|c|c|}
\hline
The sequence                &       Original RMSE, m          & RMSE with filtering, m        & \% dropped\\
\hline
tum-freiburg2-pioneer-slam2 &      0.159 $\pm$ 0.043          &       0.205 $\pm$ 0.054       &   61  \\
\hline      
tum-freiburg2-pioneer-slam3 &      0.141 $\pm$ 0.039          &       0.181 $\pm$ 0.102       &   59  \\
\hline 
\end{tabular}
\end{table}

\section{Conclusion}
This paper presents the novel filtering algorithm of laser scans from lidars, that is based on scan correspondence to each other. If an upcoming scan is similar to several previous one, then it might be dropped without loss of accuracy. The experiments were made on MIT and TUM dataset for vinySLAM, Gmapping and Cartographer. They show that it is possible to drop nearly half of the scans, and sometimes it is enough to have only one third of the total number of scans in the data sequence. 

Secondly, the algorithms of corridor detections is presented. Its accuracy varies depending on the corridor concept and might be different in specific circumstances. In MIT dataset, the corridor between office rooms is detected with this algorithm in every sequence.

Experiments show, that the suggested filtering algorithm works incomparably faster, than a scan matcher algorithm ($5.9\cdot10^{-5}$ seconds for filtering, $12.9\cdot10^{-3}$ for scan matching of vinySLAM). Therefore, the filtering process, in general, saves computational resources, because the time saved from scan matching might be used for other possible needs of a robot. Application of this filter reduces the average scan processing time by more than 40\%.

This performance is achieved by reducing the dimension of a laser scan by creation a histogram. There are several types of histograms to be created, such as the number of points on a specific distance from scanner or an average range in a specific angle of view. Then, the histograms of successive scans are compared to each other with calculating a Pearson correlation coefficient. If the correlation is high the scan should not be processed.

Finally, the paper presents formulas and suggestions for parameters of filter that depends on a robot speed. 
The future work includes updating filtering mechanism with Transferable Belief Model, that is the feature of vinySLAM - the algorithm, that was chosen for testing. Also, it is possible to apply filtering for structure-from-motion algorithms to find out if it also reduces their computational cost.

\section*{Acknowledgements}

This work was supported by the Ministry of Science and Higher Education of the Russian Federation by the Agreement N075-15-2020-933 dated 13.11.2020 on the provision of a grant in the form of subsidies from the federal budget for the implementation of state support for the establishment and development of the world-class scientific center Pavlov center Integrative physiology for medicine, high-tech healthcare, and stress-resilience technologies. Some materials and equipment has been provided by JetBrains Research.

%The citation must be used in following style: \cite{article-minimal} \cite{article-full} \cite{article-crossref} \cite{whole-journal}.

%\bibliography{xampl}
%\bibliographystyle{model3a-num-names}



%\begin{thebibliography}{}

%\bibitem{Vander}
%Van der Geer J, Hanraads JAJ, Lupton RA. The art of writing a
%scientific article. {\it J Sci Commun} 2000;{\bf 163}:51-9.
%\bibitem{Strunk}
%Strunk Jr W, White EB. {\it The elements of style}. 3rd ed. New York:
%Macmillan; 1979.
%\bibitem{Mettam}
%Mettam GR, Adams LB. How to prepare an electronic version of your article. In: Jones BS, Smith RZ, editors. {\it Introduction to the electronic age}. New York: E-%Publishing Inc; 1999. p. 281-304.

% \end{thebibliography}

\bibliographystyle{model3a-num-names}
\bibliography{paperbib}

\end{document}